# 1

# Detecting Danger: The Dendritic Cell Algorithm

Julie Greensmith[1], Uwe Aickelin[1], and Steve Cayzer[2]

[1] School of Computer Science, University of Nottingham, Jubilee Campus,
  Wollaton Road Nottingham, NG8 1BB, UK
  `[jqg, uxa]@cs.nott.ac.uk`
[2] Hewlett Packard Labs, Filton Road, Stoke Giford, Bristol, BS34 8QZ, UK
  `steve.cayzer@hp.com`

**Summary.** The Dendritic Cell Algorithm (DCA) is inspired by the function of the dendritic cells of the human immune system. In nature, dendritic cells are the intrusion detection agents of the human body, policing the tissue and organs for potential invaders in the form of pathogens. In this research, and abstract model of DC behaviour is developed and subsequently used to form an algorithm, the DCA. The abstraction process was facilitated through close collaboration with laboratory- based immunologists, who performed bespoke experiments, the results of which are used as an integral part of this algorithm. The DCA is a population based algorithm, with each agent in the system represented as an 'artificial DC'. Each DC has the ability to combine multiple data streams and can add context to data suspected as anomalous. In this chapter the abstraction process and details of the resultant algo- rithm are given. The algorithm is applied to numerous intrusion detection problems in computer security including the detection of port scans and botnets, where it has produced impressive results with relatively low rates of false positives.

## 1.1 Introduction

The Dendritic Cell Algorithm (DCA) is a biologically-inspired technique, de- veloped for the purpose of detecting intruders in computer networks. This algorithm belongs to a class of biologically inspired algorithms known as Ar- tificial Immune Systems [de Castro and Timmis, 2002]. Such algorithms use abstract models of the immune system to underpin algorithms capable of per- forming some useful computational task. The human immune system is a rich source of inspiration as it provides a high level of protection for the host body, without causing harm to the host [Coico et al., 2003].

As the name suggests, the DCA is based on a metaphor of naturally oc- curing dendritic cells (DCs), a type of cell which is native to the innate arm of the immune system. DCs are responsible for the initial detection of in- truders, including bacteria and parasites through responding to the damage caused by the invading entity. Natural DCs receive sensory input in the form of molecules which can indicate if the tissue is healthy or in distress. These cells have the ability to combine these various signals from the tissue and to produce their own output signals. The output of DCs instructs the responder cells of the immune system to deal with the source of the potential damage. DCs are excellent candidate cells for abstraction to computer security as they
are the body's own intrusion detection agents

The DCA is a multi-sensor data fusion and correlation algorithm, that can perform anomaly detection on ordered datasets, including real-time and time- series data. The signal fusion process is inspired by the interaction between DCs and their environment. In a similar manner, the DCA uses a population of agents, each representing an individual DC which can perform fusion of signal input to produce their own signal output. The assessment of the signal output of the entire DC population is used to perform correlation with 'suspect' data items. Further details of this mechanism and of the function of the DCA are presented in Section 1.4.

In this chapter the history of the development of the DCA is presented, including a brief overview of the abstract biology used to underpin the algor- tithm. This is followed by a detailed description of a generic DC based algo- rithm, including pseudocode and worked example calculations. This chapter concludes with a discussion of the applications of the algorthim to date, and application areas to which the algorithm could be applied are suggested.

## 1.2 Biological Inspiration

### 1.2.1 Danger, Death and Damage

The immune system is a decentralised, robust, complex adaptive system. It performs its function through the self-organised interaction between a diverse set of cell populations. Classically, immunology has focussed on the body's ability to discriminate between protein molecules belonging to 'self' or 'non- self', through the careful selection of cells during foetal and infant stages. This theory has underpinned the research performed in immunology since its conception by Paul Ehrlich in 1891 [Silverstein, 2005]. However, numerous problem have been uncovered with this paradigm. For example, if the immune system is tuned to respond only to non-self then why do autoimmune diseases occur, such as multiple sclerosis and rheumatoid arthritis? Why do intestines contain millions of bacteria, yet the immune system does not react against
these colonies of non-self invaders?

In 1994, immunologist Polly Matzinger controversially postulated that the immune system's objective is not to discriminate between self and nonself, but to react to signs of damage to the body. This theory is known as the Danger Theory [Matzinger, 1994]. This theory postulates that the immune system responds to the presence of molecules known as danger signals, which are released as a by-product of unplanned cell death, necrosis. When a cell undergoes necrosis, the cell degrades in a chaotic manner, producing various molecules (collectively termed 'the danger signals'), formed from the oxidation and reduction of cellular materials. Dendritic cells are sensitive to increases in the amount danger signals present in the tissue environment, causing their maturation which ultimately results in the activation of the immune system.

There are two sides to the danger theory: activation and suppression. While the presence of danger signals is sufcient to activate the immune system, the presence of a diferent class of signal can prevent an immune response. This mechanism of suppression arises as a result of apoptotic cell death, which is the normal manner in which cells are removed from the body. When a cell undergoes this process of apoptosis, it releases various signals into the environment. DCs are also sensitive to changes in concentration of this signal. DCs can combine the danger and safe signal information to decide if the tissue environment is in distress or is functioning normally. The danger theory states that the immune system will only respond when damage is indicated and is actively suppressed otherwise.

In addition to the danger theory related signals, one other class of sig- nal is processed as environmental input by DCs. These signals are termed PAMPs (pathogenic associated molecular patterns) and are a class of molecule that are expressed exclusively by micro-organisms such as bacteria. The 'in- fectious nonself' theory of immunology, developed by Janeway in the late 1980s [Janeway, 1989] states that the immune system will respond by attack- ing cells which express PAMP molecules. PAMPs are biological signatures of potential intrusion.

### 1.2.2 Introducing Dendritic Cells

DCs are the immune cells which are sensitive to the presence of danger sig- nals in the tissue. In addition to danger signals, DCs are also sensitive to two other classes of molecule namely PAMPs and 'safe' signals. PAMPs are molecules produced by microorganisms and provide a fairly definitive indi- cator of pathogenic presence. Safe signals are the opposite of danger signals, and are released as a result of controlled, planned cell death. In response to the collection of signals, the DC produces its own set of output signals - the relative concentrations of the output signals is dependent on the relative con- centrations of the input signals over time. It is the combination of external signals and current internal state which results in what is defined in this work as context.

In addition to the processing of environmental signals, DCs also collect proteins termed 'antigen'. DCs have the ability to combine the signal infor- mation with the collected antigen to provide 'context' for the classification of antigen. If the antigen are collected in an environment of mainly danger and PAMP signals, the context of the cell is 'anomalous' and all antigen collected by the cell are deemed as potential intruders. Conversely, if the environment contains mainly safe signals, then the context of the cell is 'normal' and all collected antigen are deemed as non-threatening. This theory contrasts the classical self non-self theory as the structure of the antigen proteins is not used as a basis of classification, the context is used to determine if an antigen is derived from a potential invader. The structure of the antigen is important for the subsequent response, but the processing performed by DCs involves the examination of the tissue 'context' and are unafected by the structure of the antigen.

In the natural system, this antigen-plus-context information is passed on to a class of responder cells, termed T-cells. The T-cells translate the information given to them by the local DC population. If sufcient DCs present a particular antigen to T-cells in an anomalous context, the immune system responds by eliminating any cell containing that antigen. It is noteworthy that this is a simplified description of a highly complicated immune function. For more information on the action of T-cells, please refer to a standard immunology text such as Janeway [Janeway, 2004].

The description above is a simplified description of the events which occur *in vivo*. For readers interested in the exact mechanism of DC function, refer to Lutz and Schuler [Lutz and Schuler, 2002]. In this chapter, these principles are abstracted to form a model of DC behaviour which is described in Section 1.3.

## 1.3 Abstract Model

### 1.3.1 The Approach

The DCA has been developed as part of an interdisciplinary project, known as the 'Danger Project' [Aickelin et al., 2003], which comprised a team of re- searchers including practical immunologists, computer scientists and computer security specialists. The aim of the project was to bring together state-of-the- art immunology with artificial immune systems to improve the results of such systems when applied to computer network intrusion detection. The abstract model presented in this section is the result of the collaboration between the computer scientists and immunologists. Thorough analysis of the literature as- sisted the interdisciplinary collaboration, facilitating the performance of the immunological research which contributes to the results of the abstraction process. Following this important development, key published findings from DC biology were collated.

To meet the needs of the development of the algorithm, and to further research in immunology, aspects of DC function are investigated. This in- cludes the characterisation of signals and the efects of DCs on the responder cells. Various wet-lab experiments have been performed using natural DCs to determine this necessary information [Williams et al., 2007], results of which assist in clarifying certain aspects of DC function. This research is performed following intense discussion and debate between computer scientists and immunologists, and is mutually beneficial. A diagram of the process used to develop the DCA is shown in Figure 1.1.

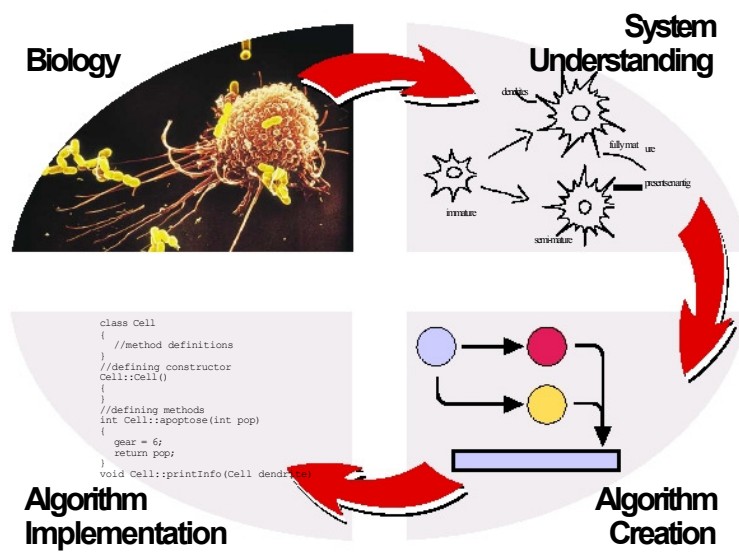

**Fig. 1.1.** A depiction of the abstraction process used in this chapter, and the rela- tionship between abstraction and immunology.

### 1.3.2 Abstract DC biology

As explained in Section 1.2, the biological function of DCs is as a natural intrusion detector. The mechanisms by which it performs this function are complex, numerous and still debated within immunology [Matzinger, 2007]. To produce an algorithm, the disparate information regarding DC biology must be combined to form an abstract model. The developed abstract model forms the basis of the DCA. Several key properties of DC biology are used to form the abstract model. These properties are *compartmentalisation, difer- entiation, antigen processing, signal processing* and *populations*.

*Compartmentalisation* provides two separate areas in which DCs perform sampling and analysis. The processing of input signals and collection of anti- gen occur in 'tissue', which is the environment monitored by DCs. Upon mat- uration DCs migrate to a processing centre, termed a lymph node. Whilst in the lymph nodes, DCs present antigen coupled with context signals, which is interpreted and translated into an immune response. In nature, this is de-signed to keep potentially deadly T-cells away from direct contact with the tissue until it is required.

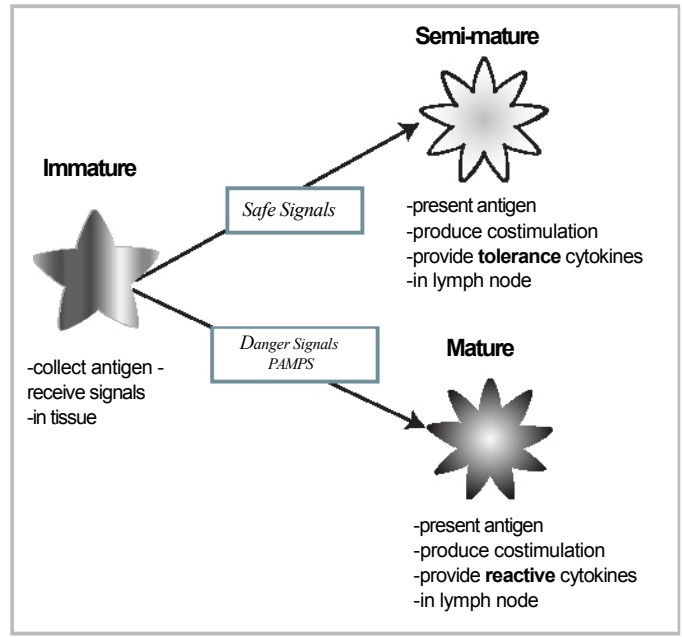

**Fig. 1.2.** An abstract model of the diferentiation of DCs, showing the transforma- tion between states and the signals responsible for the transitions. The inflammatory signal (not depicted) acts to amplify the efects of all other signals.

In this model DCs exist in one of three states, termed its state of *diferen- tiation*: immature, semi-mature and mature. Transitions to semi-mature and mature occur through the diferentiation of the immature DC. This trans- formation is initiated upon the receipt of input signals. The resultant DC state is determined through the relative proportions of input signal categories received by the immature cell. The terminal state of diferentiation dictates the context of antigen presentation where 'context' is an interpretation of the state of the signal environment. Semi-mature implies a 'safe' context and ma- ture implies a 'dangerous' context. This is a pivotal decision mechanism used by the immune system, and is the cornerstone of this abstract model.

*Antigen processing* through collection and presentation is vital to the func- tion of the system. The pattern matching of the antigen structure is not used in this model unlike previous AIS models [Balthrop et al., 2002]. The collec- tion of antigen is not responsible for the activation of the immune system although it is necessary for antigen to be sampled in order to have an en- tity to classify. This is analogous to sampling a series of 'suspects' or data to classify. The process of a immature DCs collecting multiple antigen forms the sampling mechanism used by the DCA. Each DC collects a subset of the total antigen available for sampling.

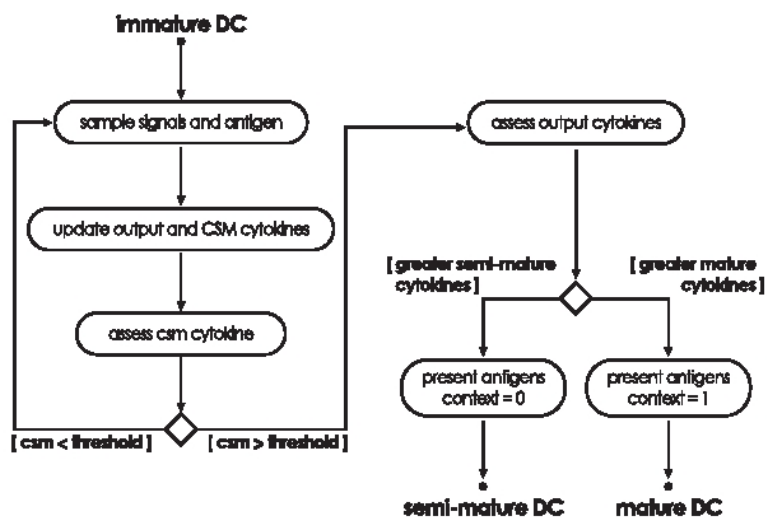

**Fig. 1.3.** A UML activity diagram representing the key features of DC biology, presented in a systemic manner. The processes on the left occur in the tissue, on the right in the lymph node.

Dendritic cells perform a type of biological *signal processing*. DCs are sen- sitive to diferences in concentration of various molecules found in their tissue environment. Safe signals are the initiators of maturation to the semi-mature state. Danger signals and PAMPs are responsible for maturation to the fully- mature state. Simultaneous receipt of signals from all classes increases the production of all three output signals, though the safe signal reduces the ex- pected amount of mature output signal generated in response to danger and PAMP signals [Williams et al., 2007]. Output signals are generated at con- centrations proportional to the input signals received.

DCs do not perform their function in isolation, residing in tissue as a pop- ulation. Each member of the population can sample antigen and signals. This multiplicity of DCs is an important aspect of the natural system. Multiple DCs are required to present multiple copies of the same antigen type in order to invoke a response from the adaptive immune system. This is an error toler- ant component of DC behaviour as it implies that a misclassification by one cell is not enough to stimulate a false positive error from the immune system. Using a population of DCs also means that diversity can be generated within the population, such as assigning each DC its own threshold values, if desired. Such diversity may also add robustness to the resultant process, presented in Figure 1.3.

| Signal | Biological Property | Abstract Property | Computational Example |
|---|---|---|---|
| PAMP | Indicator of microbial presence | Signature of likely anomaly | Error messages per second |
| Danger signals | Indicator of tissue damage | High levels indicate *potential* anomaly | Network packets per second |
| Safe signals | Indicator of healthy tissue | High levels indicate normally functioning system | Size of network packets |
| Inflammation | Indicating general tissue distress | Multiplies the all other input signals | User physically absent |

**Table 1.1.** Biological signal functions and their abstracted counterparts.

### 1.3.3 Signals and Antigen Overview

As this model is in part inspired by the danger theory, various signals drive the system. In the natural system the signals are a reflection of the state of the environment. Four categories of signal are used in this abstract model, inclusive of PAMPs, danger signals, safe signals and inflammation. The various categories of signal direct the DC population down two distinct pathways, one causing the activation of the immune system, and one responsible for generating peripheral tolerance. Upon examination of the relevant biology, it appears that DCs process all categories of signal stated above to produce their own output signals [Lutz and Schuler, 2002]. The output signals include a costimulation signal (CSM) which shows that the cell is prepared for antigen presentation and two context signals, the mature and semi-mature output signals. An overview of the names and functions of the biological signals and their abstracted counterparts is given in Table 1.1. In the forthcoming sections (1.3.4-1.3.8) all signals used in the abstract DC model are explained and rationalised individually.

### 1.3.4 PAMPs

In a biological context, PAMPs are essential products produced by microorganisms, but not produced by the host. These molecules are not unique to pathogens, but are produced by microbes, regardless of their potential pathogenicity [Medzhitov and Janeway, 2002]. PAMP molecules are a firm indicator to the innate immune system that a non-host based entity is present. Specific PAMPs bind to specific receptors on DCs (termed pattern recognition receptors) which can lead to the production two output signal molecules. These output signals are termed envision molecules (CSM) and the 'mature' output signal. Both of these chemical outputs can indicate a likely presence of a foreign entity. In this abstract model, a PAMP is interpreted as a signal which is a confident indicator of an abnormality. An increase in the strength of the PAMP input signal leads to an increase in two of three potential output signals, namely the CSM signal and the mature output signal, produced by the artificial DCs in the abstract model.

In the abstract model, PAMPs are certain indicators of an anomaly. This is based on their role *in vivo* as signatures of bacterial presence. In this research this is translated as mapping to a signature of intrusion or abnormally high rate of errors when the DCA is applied to computer security problems. For example, when applied to the detection of scanning activity, a high frequency of networking errors is translated as a high value of PAMP signal.

### 1.3.5 Danger signals

In the human immune system, danger signals are released as a result of un- planned cell death. Specifically danger signals are the by-product of cellu- lar degradation in an uncontrolled manner. The constituent components of danger signals are formed from the erratically decomposing macro-molecules normally found inside the cell, encapsulated by the cell membrane. They are indicators of damage to tissue, which the immune system is trying to protect. In a similar manner to PAMPs the receipt of danger signals by a DC also causes diferentiation to the fully mature state. However, the resultant efect on DCs through danger signals is less than that of PAMPs. This means that a higher concentration of danger signal molecules are needed in order to elicit a response of the same magnitude as with a similar concentration of PAMPs, where concentration is the number of molecules of signal per unit volume.

Within the context of the abstract model, danger signals are indicators of abnormality, but have a lower value of confidence than associated with the PAMP signal. The receipt of danger signals also increases the amount of CSMs and mature output signals produced by the DC. The receipt of danger signals causes the presentation of antigen in a *dangerous context*. This can ultimately lead to the activation of the adaptive immune system. In a computational context, for example to detect scanning activity on a computer network, the danger signal can be derived from the rate of sent/recieved network packets per second. A high rate of sending of packets may be indicative of an anomaly at high levels, but at low levels is likely to indicate normal system function.

### 1.3.6 Safe signals

Within the natural immune system, certain signals are released as a result of healthy tissue cell function. This form of cell death is termed apoptosis - the signals of which are collectively termed 'safe signals' in this work. The receipt of safe signals by a DC results in the production of CSMs in a similar manner to the increase caused by PAMPs and danger signals. In addition the 'semi- mature' output signal is produced as a result of the presence of safe signals in the tissue. The production of the semi-mature output signal indicates that antigen collected by this DC was found in a normal, healthy tissue context. Tolerance is generated to antigen presented in this context.

The secondary efect of safe signals, is their influence on the production of the mature output signal. In the situation where tissue contains cells under- going both apoptosis and necrosis, the receipt of safe signals suppresses the production of the mature output signal in response to the danger and PAMP signals present in the tissue [Williams et al., 2007]. This appears to be one of many regulatory mechanism provided by the immune system to prevent the generation of false positives. This is a key mechanism of suppression of the response to antigen not directly linked to a pathogen. The balance between safe and danger signals and the resultant effects on the production of the ma- ture output signal is incorporated in the signal processing mechanism. The incorporation of this mechanism is significant for the danger project as its use was facilitated by the close collaboration achieved with the team of laboratory based immunologists.

Within this abstract model, input signals which indicate normality are termed 'safe signals'. This signal is interpreted as data which indicates normal system/data behaviour and a high level of this signal will increase the output signal value for the 'semi-mature signal'. In-line with the biological efect of this signal, subsequent receipt of a high safe signal value will reduce the cumulative value of the 'mature' output signal, incremented by the receipt of either PAMPs or danger signals. The interaction between these signals is shown in Figure 1.5.

In a computational context, for example to detect scanning activity on a computer network, the safe signal is an indicator of normal machine behaviour, which can also be derived from the rate of sent/recieved network packets per second. In previous work it is identified that scanning activity produces highly 'regular' and small network packet sizes. Therefore the safe signal value is produced in proportion to the average packet size, with a high safe signal value created if the average packet size is sufciently larger than the expected size.

### 1.3.7 Inflammation

As shown in [Sporri and Caetano, 2005], the presence of inflammatory sig- nals in human tissue are insufcient to initiate maturation of an immature DC. However, the presence of inflammation not only implies the presence of inflammatory cytokines (cytokines are biological signals which act as messen- ger molecules between cells) but also that the temperature is increased in the affected tissue. Additionally, the rates of reaction are increased because of this increasing heat, plus inflammatory cytokines initiate the process of dilating blood vessels, recruiting an increased number of cells to the tissue area under distress.

A variant of this concept is employed in the abstract model, where inflam- mation has the efect of amplifying the other three categories of input signal, inclusive of safe signals. The resultant efect of the amplification is an increase in the artificial DC's output signals. An increase in inflammation implies that the rate of DC migration will increase, as the magnitude of the CSMs pro- duced by the DC will occur over a shorter duration, and hence resulting in a shortened DC life span in the tissue compartment. It is important to stress however, that the presence of inflammatory signals alone is insufcient to instruct the immune system how to behave appropriately.

### 1.3.8 Output signals

From examination of the biological literature, it is evident that DCs produce a set of output signals as a result of exposure to the environmental input signals experienced in the tissue. By process of abstraction, three signals in particular are selected to be the output signals of the DCs:

1. CSM output: limits the lifespan of a DC, through being assessed against a *migration threshold*
2. Semi-mature: output incremented in response to safe signals
3. Mature: output incremented in response to PAMP and danger signals; reduced in response to safe signals

In the natural system DC CSM production is combined with production of another receptor which attracts the DC to the lymph node for antigen presentation, where the DCs present their antigen to a responder cell. This mechanism is complicated and is abstracted to a simpler version for use within an algorithm. In the abstract model an increased amount of CSMs increases the probability of a DC leaving the tissue and entering the lymph node for analysis. This is abstracted into a model through the assignment of a migration threshold, which is described in detail in Section 1.4. In the abstract model, if this threshold is exceeded, the state of the cell changes from immature to either semi-mature or mature. The cell then enters the *antigen presentation stage* where its context is assessed.

In nature, the presence or absence of these two chemicals controls the response of the responder cells. In this presented model, these responder cells do not feature, and therefore the information provided through the use of these context signals is used in a diferent manner. The context of the DC in the abstract model is controlled by the relative proportions of the semi-mature signal to the mature signal. The DCs context is assigned by whichever of the two output signals is greater upon presentation of antigen by the DC. A larger value of semi-mature signal implies the presented antigen was collected in a primarily 'normal' context whereas a larger value of the mature output

| Biological | Abstract |
|---|---|
| PAMP | PAMP |
| necrotic products | danger signals |
| apoptotic cytokines | safe signals |
| inflammatory cytokines | inflammation |

**Table 1.2.** Biological and abstract computational terms for the input signals.

signal would imply that the presented antigen was collected in a potentially 'anomalous' context.

### 1.3.9 Signal Summary

In Table 1.2 the various synonyms for the various terms at diferent levels of abstraction is given, from the actual biological terms to the terms used in the general model of a DC based algorithm. A state chart showing the influence of the various signals and the corresponding output signals is presented in Figure 1.4, where IL-12 and IL-10 are the mature and semi-mature output signals respectively.

### 1.3.10 Accounting for Potency: Signal Processing

The actual mechanisms of internal DC signal processing are vastly complex and are termed signal transduction mechanisms. For the purpose of the ab- stract model and resultant algorithm a simplified version of signal processing can be implemented without compromising the underlying metaphor. An ab- stracted model of signal transduction is developed, which accounts for the magnitude of responses, but does not involve the intricacies of a signaling network. This interaction is simplified to a weighted sum equation, which is performed for the transformation of input signals to output signals. A rep- resentation of this process is shown in Figure 1.5. The influence of a signal on a cell, the potency, is translated as the weight value given to each signal, and efcacy represented as either a positive or negative weight value. In the system presented in this chapter, the weight values given above are used as an integral part of the system, and it is repeatedly shown that these values suit the chosen applications. However, these weights are given as a guideline - other values may be more suitable for diferent applications. This may become apparent as the DCA is applied to a more diverse set of applications.

### 1.3.11 Abstract Antigen

The combination of signals provides the basis of classification which can be used for the purpose of anomaly detection. The processing of signals would be sufcient to indicate if the tissue is currently in distress or under attack, but it would not yield any information regarding the originator of the anomaly, namely the culprit responsible. Antigen is required in order to link the evi- dence of the changing behaviour of tissue with the culprits which may have caused this change in behaviour. Antigen is necessary: it is the data that is to be classified, with the basis of classification derived not from the structure of this antigen but from the relative proportions of the three categories of input signal, processed across a population of DCs.

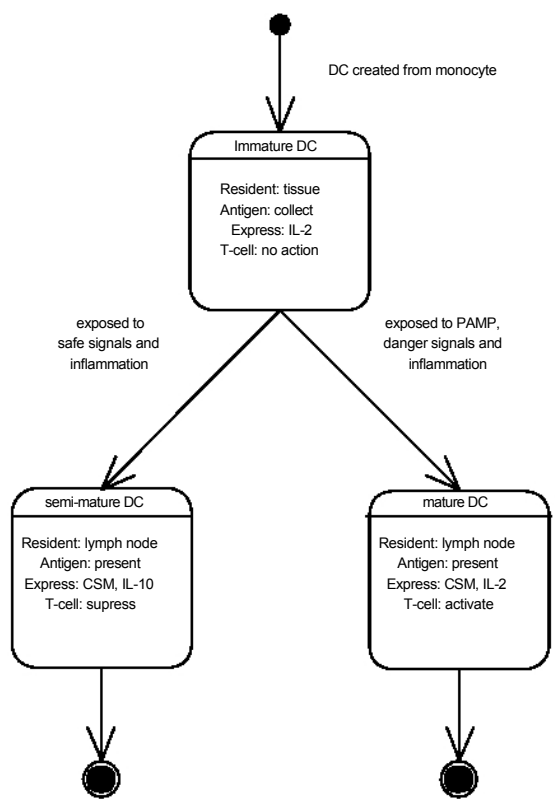

**Fig. 1.4.** A state chart showing various DC states and the featured input and output signals, where responder cells are termed T-cells.

It is important to note that a single antigen of a specific structure will not be sufcient to elicit any response from the immune system. Concentrations of antigens with identical structures are found in tissue and processed by the DCs. In selecting suitable data, either multiple items with the same struc- ture should be used, forming an *antigen type*. Aggregate sampling of multiple antigens is a key property of the system, and may provide some robustness and tolerance against rogue signal processing of a small number of DCs. In this abstraction no processing of antigen is performed as the focus is on the treatment of the diferent categories of input signal.

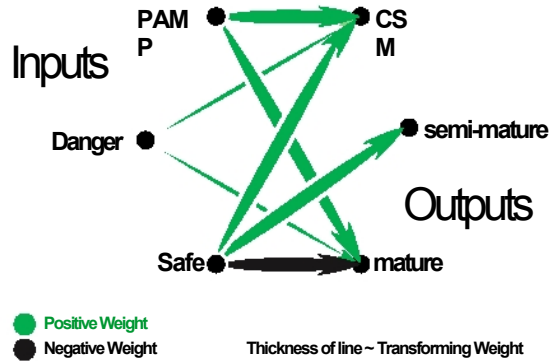

**Fig. 1.5.** An abstract model of DC signal processing. The inflammatory signal (not pictured) acts as a general amplification signal.

### 1.3.12 Assumptions and Simplifications

As part of this abstraction process various assumptions and simplifications are made, as the purpose of this process is to derive a feasible algorithm, not produce a realistic simulator of DC biology. It is assumed that no other type of immune cells are required for this algorithm to function. Unlike the approach of [Twycross, 2007], DCs in the DCA function in isolation and the T-cell component is replaced with a statistical technique. This is possible as system changing responses do not form part of this model. It is also assumed that no inter-cell communication occurs and that individual DCs do not communicate with one another. This can be assumed as no adaptation is present in this system.

It is assumed that four signal categories exist, and that the DC does not respond to any other signal. Of course, DCs express a plethora of receptors for various molecules. In this abstraction only the molecules responsible for immune activation are used. In a similar manner it is assumed that three output signals are produced. It is also assumed that DCs are impervious to unexpected death, unlike in the human immune system. In this model a single tissue compartment is used.

The above assumptions are used to make the abstraction clearer and the resultant algorithm simpler to understand. There are also various assumptions in this abstraction which are made due to the lack of understanding of natural DCs within immunology. In this abstraction it is assumed that each DC has a fixed size capacity for antigen storage. This is assumed as there is no biological data available to confirm the antigen capacity of DCs. In a similar manner, it is unknown which agent is responsible for limiting the sampling period of the DCs within the tissue. In this abstraction, measurement of CSMs against a migration threshold determines the duration of the DCs life span. As the objective of this work is to produce an algorithm, not n accurate simulation, it is acceptable to make such assumptions, provided they are useful in leading to a feasible algorithm.

## 1.4 The Dendritic Cell Algorithm

### 1.4.1 Algorithm Overview

The development of an abstract model of DC behaviour is one step in the development of a danger theory inspired intrusion detection system. To trans- form the abstract model of DC biology into an immune-inspired algorithm, it must first be formalised into the structure of a generic algorithm and into a series of logical processes. It must also be expressed appropriately so that the DCA can be implemented feasibly. A generic form of the algorithm is given in this section. For further details of the algorithm and for information regarding its implementation as a real-time anomaly detection system, please refer to Greensmith *et al*. [Greensmith et al., 2008].

The purpose of a DC algorithm is to correlate disparate data-streams in the form of antigen and signals. The DCA is not a classification algorithm, but shares properties with certain filtering techniques. It provides information representing how anomalous a group of antigen is, not simply if a data item is anomalous or not. This is achieved through the generation of an anomaly coef- ficient value, termed the *MCAV - mature context antigen value*. The labelling of antigen data with a MCAV coefcient is performed through correlating a time-series of input signals with a group of antigen. The signals used are pre- normalised and pre-categorised data sources, which reflect the behaviour of the system being monitored. The signal categorisation is based on the four signal model, based on PAMP, danger, safe signals and inflammation. The co-occurrence of antigen and high/low signal values forms the basis of cate- gorisation for the antigen data.

This overview, whilst technically correct, is still somewhat abstract. To ce- ment the ideas which form the DCA, a generic representation of the algorithm is presented. A formal description of the algorithm and details of it's imple- mentation are presented in [Greensmith et al., 2008] and [Greensmith, 2007]. To further elaborate on the workings of a DC based algorithm, each key com- ponent is described in turn. The primary components of a DC based algorithm
are as follows:

1. Individual DCs with the capability to perform multi-signal processing
2. Antigen collection and presentation

---

**Algorithm 1**: Pseudocode of the functioning of a generic DC object.

    **input** : signals from all categories and antigen
    **output**: antigen plus context values

    initialiseDC;

    **while** *CSM output signal < migration Threshold* **do**
        | get antigen;
        | store antigen;
        | get signals;
        | calculate interim output signals;
        | update cumulative output signals;
    **end**
    cell location update to lymph node;

    **if** *semi-mature output > mature output* **then**
        | cell context is assigned as 0 ;
    **else**
        | cell context is assigned as 1;
    **end**
    kill cell;
    replace cell in population;

---

3. Sampling behaviour and state changes
4. A population of DCs and their interactions with signals and antigen
5. Incoming signals and antigen, with signals pre-categorised as PAMP, danger, safe or inflammation
6. Multiple antigen presentation and analysis using 'types' of antigen
7. Generation of anomaly coefcient for various diferent types of antigen

The DCA is a population based algorithm, with the population consisting of a set of interacting objects, each representing one cell. Each cell in the population has a set of instructions which is followed each time a cell is updated. The control of the frequency and nature of cell updates is specific to the implementation and will be described in the second half of this chap- ter. Each DC object within the population performs its own antigen sampling and signal collection. Diversity is generated within the DC population by ini- tiation of migration of the DCs at diferent time points i.e. the cessation of data sampling. This creates a variable time window efect throughout the DC population, which may add some robustness to the system.

### 1.4.2 An Individual DC

As aforementioned, each DC in the system is represented by an object, capa- ble of executing its own behavioural instructions. DCs process input signals to form a set of cumulatively updated output signals in addition to the collection of antigen throughout the duration of the sampling stage. Each DC can exist in one of three states at any point in time. These states are namely imma- ture, semi-mature or mature. The relationship between these various states is shown in the UML state chart in Figure 1.6. However, the diferences in the semi-mature and mature state is controlled by a single variable, determined by the relative diferences between two output signals produced by the DCs. The initiation of the state change from immature to either mature or semi-mature is facilitated not by the collection of antigen, but by sufcient exposure to sig- nals. This exposure is limited by the assigned *migration threshold*. Pseudocode of a generic DC object is given in Algorithm 1.

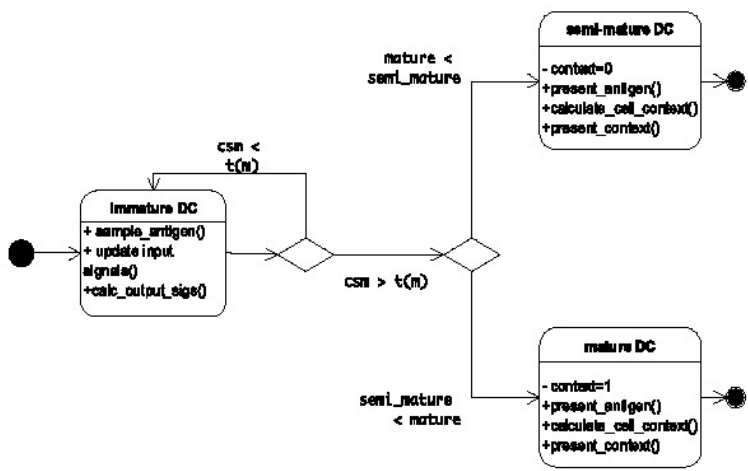

**Fig. 1.6.** DC state transition overview diagram, showing the diferent possible states over a DC life span, where t(m) represents the migration threshold of the cell and CSM is the costimulatory output signal.

Whilst in the immature state, the DC has three functions, which are per- formed each time a single DC is updated:

1. *Sample antigen:* the DC collects antigen from an external source (in this case, from the 'tissue') and places the antigen in its own antigen storage data structure.
2. *Update input signals:* the DC collects values of all input signals present in the signal storage area
3. *Calculate interim output signals:* at each iteration each DC calculates three temporary output signal values from the received input signals, with the output values then added to form the cell's cumulative output signals.

Signal processing occurs within DCs of the immature state. The signal pro- cessing performed is suggested to be in the form of a weighted sum equation, bypassing the modelling of any biologically realistic gene regulatory network or signal transduction mechanism. A simple weighted sum equation is used in order to reduce any additional computational overheads, with the intended DCA application being real-time anomaly detection. In the generic algorithm, the only crucial component of this procedure is the ability of the end user to map raw input data to one of the four categories of input signal (PAMP, danger, safe and inflammation). The general form of the signal processing equation is shown in Equation 1.1, where $P_w$ are the PAMP related weights, $D_w$ for danger signals etc.

| Signal | PAMP | Danger | Safe |
|---|---|---|---|
| CSM | $W1$ | $\frac{W1}{2}$ | $W1*1.5$ |
| Semi-mature | 0 | 0 | 1 |
| Mature | $W2$ | $\frac{W2}{2}$ | $-W2*1.5$ |

**Table 1.3.** Derivation and interrelationship between weights in the signal processing equation, where the values of the PAMP weights are used to create the all other weights relative to the PAMP weight. $W1$ is the the weight to transform the PAMP signal to the CSM output signal and $W2$ is the weight to transform the PAMP signal to the mature output signal.

$$Output = (\sum(P_n * P_w) + \sum(D_n * D_w) + \sum(S_n * S_w)) * (1+I) \qquad (1.1)$$

In the generic form of the signal processing equation (Equation 1.1) $P_n$, $D_n$ and $S_n$ are the input signal value of category PAMP (P), danger (D) or safe (S) for all signals (n) of that category, assuming that there are multiple signals per category. In this equation, $I$ represents the inflammation signal. This sum is repeated three times, once per output signal. This is to calculate the interim output signal values for the CSM output, the semi-mature output and mature output signals. These values are cumulatively summed over time.

The weights used in this signal processing procedure are derived empiri- cally from immunological data, generated for the purpose of the model devel- opment [Williams et al., 2007]. From past experience these are combinations that have worked well, shown through sensitivity analysis to work for the chosen applications - though they are not fundamental to the algorithm. The actual values used for the weights can be user defined, though the relative values determined empirically are kept constant. The relative weight values are presented in Table 1.3. These signals are used to assess the state of the DC upon termination of the sampling phase of a DCs life span. The three output signals of a DC perform two roles, to determine if an antigen type is anomalous and to limit the time spent sampling data. A summary of the three output signals and their function is given in Table 1.4, where:

1. To limit the life span/sampling duration of an individual cell
2. To assess which terminal DC state should be reached i.e. semi-mature versus mature.

| Output signal | Function |
| --- | --- |
| Costimulatory signal | assessed against a threshold to limit the duration of DC signal and antigen sampling, based on a migration threshold |
| Semi-mature signal | terminal state to semi-mature if greater than resultant mature signal value |
| Mature signal | terminal state to mature if greater than resultant semi-mature signal value |

**Table 1.4.** Table of cumulative output signals and their associated implications for the DCA

Within the Danger Project, the word *context* is used extensively. The word context refers to the circumstances in which an event occurs. Context means a representation of the signal circumstances in which an antigen is processed. The context used to categorise antigen is not achieved with one DC for one antigen, but the aggregate total of contexts across a population of DCs and a set of antigen. Nevertheless, each member of the DC population is assigned a context upon its state transition from immature to a matured state. Each DC makes a binary choice, as an individual cell can only be either mature or semi-mature, but not both.

Diversity and feedback in the DC population is maintained through the use of variable migration thresholds. This concept is touched upon in Section 1.3, but what implications does it actually have for the algorithm, and what exactly is a variable migration threshold? The natural mechanism of DC mi- gration is complex and not particularly well understood, involving the up and down regulation of many interacting molecules. Instead of using a model of what is ascertainable from the natural system, a surrogate mechanism which shows similar end results is implemented.

In this algorithm multiple DCs are used to form a population, each sam- pling a set of signals within a given time window'. Each DC in the population is assigned a 'migration threshold value' upon its creation. Following the up- date of the cumulative output signals, a DC compares the value it contains for CSMs with the value it is assigned as its migration threshold. If the value of CSM exceeds the value of the migration threshold the DC is removed from the sampling area and its life span is terminated upon analysis in the 'lymph node' area, which is a diferent compartment than tissue.

Each member of the DC population is randomly assigned a migration threshold upon its creation. The range of the random threshold is a user definable parameter, with this range being applicable to the whole DC popu- lation. Previous experience with the DCA, the median point about which the migration thresholds are assigned equates to a DC sampling for two iterations when the signal strengths are half the expected total input signal maximum. This process discounts the use of inflammation in this derivation. Addition- ally the range of the random assignment is ± 50% of the median value of a uniform distribution. A derivation of this is shown in Equation 1.2. In this

---

**Algorithm 2**: Context assessment for a single DC.

**input** : semi-mature and mature cumulative output signals
**output**: collected antigen and cell context

**if** *semi-mature output > mature output* **then**
| cell context is assigned as 0 ;
**else**
| cell context is assigned as 1;
**end**
print collected antigen plus cell context

---

equation $max_p$ is the maximum observed level of PAMP signal, and $weight_{pc}$ is the corresponding transforming weight from **P**AMP to **C**SM output signal.
In a similar manner $max_d$ and $max_s$ and $weight_{dc}$ $weight_{sc}$ are equivalent values for danger signal and safe signal. Inflammation is not included in this derivation.

$$t_{median} = 0.5 * ((max_p * weight_{pc}) + (max_d * weight_{dc}) + (max_s * weight_{sc})) \tag{1.2}$$

The net result of this is that diferent members of the DC population 'experience' diferent sets of signals across a time window. If the input signals are kept constant, this implies that members of the population with low values of migration threshold present antigen more frequently, and therefore produce a tighter coupling between current signals and current antigen. Conversely, DCs with a larger migration threshold may sample for a longer duration, producing a more relaxed coupling between potentially collected signal and context. Having a diverse population, whom all sample diferent total sets of signals, is a positive feature of this algorithm, demonstrated through results presented in [Greensmith et al., 2008].

Once the cell has migrated, its role is to then present the antigen and output signals it has collected throughout its life span. As part of this pro- cess the kinds of signal it was exposed to over its life span are assessed and transformed into a binary value - this is termed the DC context. This can be achieved through a simple comparison between the remaining two out- puts signals, which are resultant cumulative values. These two values (semi and mature output signals) are compared directly with each other using the relationship described in Algorithm 2.

The context is vital to assign any collected antigen with the context in which the cell performed its collection. Another important feature of the al- gorithm is that each DC can sample multiple antigens per iteration and can store these antigens (up to a certain capacity) internally for presentation upon maturation.

To summarise, each DC has the ability to process and collect signals and antigen. Through the generation of cumulative output signals, the DC forms a cell context which is used to perform anomaly detection. in the assessment of antigen. The life-span of the DC is controlled by a threshold, termed the migration threshold, which is randomly assigned to each DC in the population (within a given range). Upon migration the cumulative output signals are assessed and the greater of semi or mature output signal becomes the cell context. This cell context is used to label all antigen collected by the DC with the derived context value of 1 or 0. This information is ultimately used in the generation of an anomaly coefcient.

### 1.4.3 Populations, Tissue And Assessment - The Macroscopic Level

As mentioned the DCA is a population based algorithm, based on an agent-like system of artificially created cells whom interact with an artificially created environment. This consists of a tissue compartment and a lymph node com- partment. In the tissue compartment signals and antigen are stored for use by the DC population. DCs are transferred to the lymph node compartment for analysis upon migration. It is in the lymph node where the antigen plus context values are logged for analysis.

The interaction between cells and environment - termed here as tissue - is crucial, and drives the system. From a DC's perspective the enviroment/tissue is what it can sense. In the case of natural DCs, they sense the world around them through activation or deactivation of receptors found on the surface. Indeed the DCs outlined in the section above have a similar system of being able to sense the signal data present in the tissue and to respond through the generation of output signals.

In addition to sensing signals, DCs also interact with antigen. This is performed through the transfer of antigen from its store in the tissue com- partment to the internal storage for antigen within the sampling DC. For use in a DC based algorithm, the environment for a DC in the sampling popu- lation consist only of signals and antigen. Therefore, in a generic DC-based algorithm, tissue is comprised of signals and antigen as this is what the cell population can respond to and process.

It is proposed that the updates of antigen, signals and cells are performed independently. The dictated timing of when entities are updated is left to the user. In the real-time implementations described in this thesis, cells are updated once per second. In the implemented system, signals are also updated at a rate of once per second, with antigen updated as soon as the data becomes available. The rate of update is dependent upon the requirements of the user and the nature of the input data and application. The exact nature of the update mechanisms are not specific to the algorithm, it can be up to the user, or dictated by the nature of the data processed by the algorithm. However, it is noteworthy that each of the three updates need not occur simultaneously: this temporal correlation between asynchronously arriving data is performed by the processing of the cells themselves.

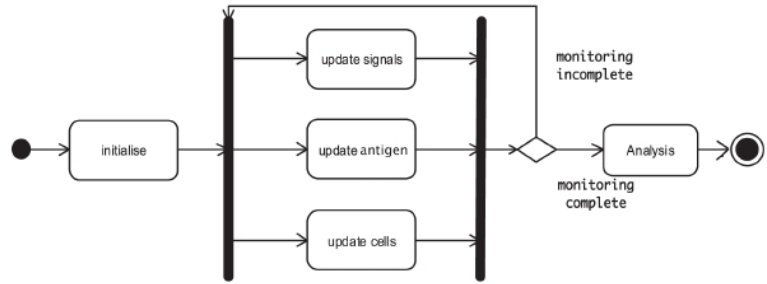

**Fig. 1.7.** A UML overview of the processes at the tissue level of the program, showing the asynchronous update of cells, signals and antigen. It also shows the two main stages of update and initialisation and subsequent analysis.

The population dynamics are used to perform the actual anomaly detec- tion. The ultimate classification of a particular type of antigen is derived not from a single DC, but from an aggregate analysis produced across the DC population over the duration of an experiment.

The derived value for the cell context is assigned to each antigen (if indeed any) collected by the assessed DC. This information is used to derive the MCAV anomaly coefcient for a particular type of antigen. This relies on the fact that during their time as sampling entities, the DCs sample both antigen and signals. This is also dependent upon the use of *antigen types*. This means that the input antigen are not unique in value, but belong to a population in themselves. Numerous experiments in this thesis, the ID value of a running program is used to form antigen, with each antigen generated every time the program sends an instruction to the low level system. Therefore a population of antigen is used, linked to the activity of the program, and all bearing the same ID number.

Each DC can sample multiple antigens per iteration and can store a fixed maximum amount of antigen within whilst sampling signals. It is the consen- sus value for an entire antigen type which gives rise to the anomaly detection within this algorithm. The MCAV is mean value of context per antigen type. Pseudocode for the generation of the MCAV is given in Algorithm 3. The closer the MCAV is to one, the more likely it is that the majority of the anti- gen existed in the tissue at the same time as a set of signals. This is similar to the principle of guilt by association, which has a temporal basis. If more than one tissue compartment were used this association would also be spatial. The 'cause and efect' means of classification is facilitated by the temporal correla- tion produced through the use of DCs whom sample signals and antigen over diferent durations.

**Algorithm 3**: The generation of MCAV coefcients for each antigen type sampled by the DC Algorithm.

**input** : total list of antigen plus context values per experiment
**output**: MCAV coefcient per antigen type

**for** *all antigen in total list* **do**
    increment antigen count for this antigen type;
    **if** *antigen context equals* 1 **then**
        increment antigen type mature count;
    **end**
**end**
**for** *all antigen types* **do**
    MCAV of antigen type = mature count / antigen count;
**end**

### 1.4.4 Generic DC Algorithm Summary

An overview of the DCS is presented in Figure 1.8. In Section 1.4 a generic description of the algorithm is presented, outlining its key features and mecha- nisms for processing data, filtering and detecting anomalous antigen. At a cell level, the DC is a signal processing unit, who makes a binary (yes/no) decision as to whether the antigen it has collected during its life span was collected un- der anomalous conditions. At a population level, the greater DC population is used to perform anomaly detection based on the consensus opinion of the col- lection of cells. This behaviour produces a robust method of detection through the incorporation of multiple antigen and signal sampling across a population of artificial cells all with variable life spans. This forms a filter-based corre- lation algorithm which includes a 'time window' efect which reduces false positive errors [Greensmith, 2007].

## 1.5 Applications: Past and Present

The DCA is designed with the objective of its ultimate application to problems in network intrusion detection, through reducing the high rates of false posi- tives previously seen with anomaly detection systems [Aickelin et al., 2004]. While the DCA has been applied to such problems [Greensmith et al., 2006], it has also enjoyed some preliminary successes in sensor networks and mobile robotics.

    Early work with the algorithm involved its application to a standard ma- chine learning data set [Greensmith et al., 2005], where it was shown that the algorithm can process classification data, but is sensitive to the data order. Once the algorithm was deemed feasible through its application to the ma- chine learning dataset, the DCA has also been applied to the detection of port scans and scanning based activity [Greensmith et al., 2008], which produced

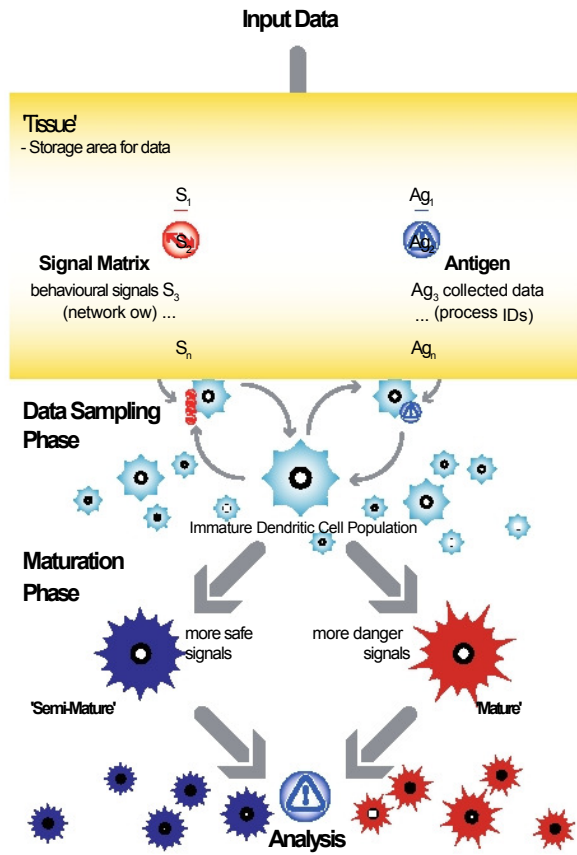

**Fig. 1.8.** Illustration of the DCA showing data input, continuous sampling, the maturation process and aggregate analysis.

high rates of true positives and low rates of false positives. In the case of the port scan experiments, signals are taken as behavioural attributes and system calls are used to form antigen. This research is ongoing and now encompasses the detection of other forms of malicious mobile code, such as botnets and scanning worms.

In addition to standard network anomaly detection tasks, Kim *et al.* [Kim et al., 2006] produced an implementation of the DCA for detecting mis-behaviour in sensor networks. The signals are based on the behaviour of packet sending and is used to determine which nodes in the network are potentially under attack. The use of the DCA in this scenario produced satisfactory re- sults. In conclusion, this problem is suitable for use with in the DCA as data fusion from disparate sources is required to perform detection.

Oates *et al.* [Oates et al., 2007] have applied the DCA to object detec- tion using mobile robots. The DCA is used to classify specific object based on combining data from various robot sensors in real-time. As part of this research, theoretical analysis of the algorithm is being performed to assist in its application to difcult robotic problems. This research indicates that the DCA is a suitable algorithm for applications in mobile robotics.

As the DCA can analyse time-dependent data in real-time, there are nu- merous areas to which the algorithm could be applied, both within computer network intrusion detection and in other more general scientific applications. For example, it may be useful in the prediction of earthquakes, by looking for 'danger' in the form of seismic activity and correlate this information with lo- cation, encoding antigen. Similar signal/location correlating problems such as the analysis of radio anomalies in space and the analysis of real-time medical data may be potential applications areas for the DCA.

## 1.6 Conclusions

In this chapter the Dendritic Cell Algorithm is presented as an immune- inspired algorithm. This algorithm is based on an abstract model of the biolog- ical dendritic cells (DCs), which are a key decision making cell of the human immune system. The abstract model presented in this chapter shows the key properties of the natural system, and such properties are presented to form a model. From this model a generic DC based algorithm is presented. This algo- rithm forms the DCA, and is capable of performing multi-sensor data fusion on the input signals, combined with a correlation component, linking signals to antigen data. The process by which the signals are used and combined is detailed, in combination with a description of the behaviour for each artificial cell within the algorithm.

The DCA has enjoyed success so far in its application to the detection of port scans, and is shown in the related work to be a robust and decentralised algorithm. The key to the robustness lies in the 'time-window efect', where diferent members of the population sample input data across diferent du- rations. This efect is thought decrease the number of false positive results produced by the algorithm.

Future developments with the DCA include the addition of a 'responder cell' component, to calculate the MCAV anomaly coefcient dynamically. This would potentially increase the sensitivity of the system. Understanding the exact workings of the DCA is a non-trivial task. So far the majority of its characterisation has been performed empirically, through sensitivity analysis and parameter modification. However, in future more theoretical approach to its analysis will be taken, through the use of various theoretical tools such as constraint satisfaction. Perhaps through the performance of this analysis it can be shown exactly why this algorithm produces the good rates of detection in a robust manner.

## References

[Aickelin et al., 2003] Aickelin, U., Bentley, P., Cayzer, S., Kim, J., and McLeod, J. (2003). Danger theory: The link between AIS and IDS. In *Proc. of the 2nd International Conference on Artificial Immune Systems (ICARIS), LNCS 2787*, pages 147-155. Springer-Verlag.

[Aickelin et al., 2004] Aickelin, U., Greensmith, J., and Twycross, J. (2004). Immune system approaches to intrusion detection - a review. In *Proc. of the 3rd International Conference on Artificial Immune Systems (ICARIS), LNCS 3239*, pages 316-329.

[Balthrop et al., 2002] Balthrop, J., Esponda, F., Forrest, S., and Glickman, M. (2002). Coverage and generaliszation in an artificial immune system. In *Proc. of the Genetic and Evolutionary Computation Conference (GECCO)*, pages 3-10.

[Coico et al., 2003] Coico, R., Sunshine, G., and Benjamini, E. (2003). *Immunology: A Short Course*. Wiley-Liss.

[de Castro and Timmis, 2002] de Castro, L. and Timmis, J. (2002). *Artificial Immune Systems: A New Computational Approach*. Springer-Verlag, London. UK.

[Greensmith, 2007] Greensmith, J. (2007). *The Dendritic Cell Algorithm*. PhD thesis, School of Computer Science, University Of Nottingham.

[Greensmith et al., 2005] Greensmith, J., Aickelin, U., and Cayzer, S. (2005). Introducing Dendritic Cells as a novel immune-inspired algorithm for anomaly detec- tion. In *Proc. of the 4th International Conference on Artificial Immune Systems (ICARIS), LNCS 3627*, pages 153-167. Springer-Verlag.

[Greensmith et al., 2008] Greensmith, J., Aickelin, U., and Tedesco, G. (2008). Information fusion for anomaly detection with the dca. *Information Fusion*, tbc(tbc):tbc.

[Greensmith et al., 2006] Greensmith, J., Aickelin, U., and Twycross, J. (2006). Articulation and clarification of the dendritic cell algorithm. In *Proc. of the 5th International Conference on Artificial Immune Systems (ICARIS), LNCS 4163*, pages 404-417.

[Janeway, 1989] Janeway, C. (1989). Approaching the asymptote? Evolution and revolution in immunology. *Cold Spring Harbor Symposium on Quant Biology*, 1:1-13.

[Janeway, 2004] Janeway, C. (2004). *Immunobiology*. Garland Science, 4th edition. [Kim et al., 2006] Kim, J., Bentley, P., Wallenta, C., Ahmed, M., and Hailes, S. (2006). Danger is ubiquitous: Detecting malicious activities in sensor networks using the dendritic cell algorithm. In *Proc. of the 5th International Conference on Artificial Immune Systems (ICARIS), LNCS 4163*, pages 390-403.

[Lutz and Schuler, 2002] Lutz, M. and Schuler, G. (2002). Immature, semi-mature and fully mature dendritic cells: which signals induce tolerance or immunity? *Trends in Immunology*, 23(9):991-1045.

[Matzinger, 1994] Matzinger, P. (1994). Tolerance, danger and the extended family. *Annual Reviews in Immunology*, 12:991-1045.

[Matzinger, 2007] Matzinger, P. (2007). Friendly and dangerous signals: is the tissue in control? *Nature Immunology*, 8(1):11-13.

[Medzhitov and Janeway, 2002] Medzhitov, R. and Janeway, C. (2002). Decoding the patterns of self and nonself by the innate immune system. *Science*, 296:298- 300.

[Oates et al., 2007] Oates, R., Greensmith, J., Aickelin, U., Garibaldi, J., and Kendall, G. (2007). The application of a dendritic cell algorithm to a robotic classifier. In *Proc. of the 6th International Conference on Artificial Immune Sys- tems (ICARIS), LNCS 4628*, pages 204-215.

[Silverstein, 2005] Silverstein, A. (2005). Paul Ehrlich, archives and the history of immunology. *Nature Immunology*, 6(7):639-639.

[Sporri and Caetano, 2005] Sporri, R. and Caetano, C. (2005). Inflammatory mediators are insufcient for full dendritic cell activation and promote expansion of cd4+ t cell populations lacking helper function. *Nature Immunology*, 6(2):163-170.

[Twycross, 2007] Twycross, J. (2007). *Integrated Innate and Adaptive Artificial Immune Systems Applied to Process Anomaly Detection*. PhD thesis, University Of Nottingham.

[Williams et al., 2007] Williams, C., Harry, R., and McLeod, J. (2007). Mechanisms of apoptosis induced DC suppression. *Submitted to the Journal of Immunology*.